\pdfoutput=1

\documentclass[11pt]{article}

\usepackage{acl}

\usepackage{times}
\usepackage{latexsym}

\usepackage[T1]{fontenc}

\usepackage[utf8]{inputenc}

\usepackage{microtype}

\usepackage{inconsolata}

\usepackage{graphicx}
\usepackage{inconsolata}
\usepackage{graphicx}
\usepackage{amsmath}
\usepackage{adjustbox}
\usepackage{multirow}
\usepackage{svg}
\usepackage[normalem]{ulem}
\useunder{\uline}{\ul}{}
\usepackage{makecell}
\usepackage{tablefootnote}

\usepackage{xcolor}

\definecolor{colorhigh}{HTML}{9900FF}
\definecolor{colorlow}{HTML}{660000}

\title{Building an Efficient Multilingual Non-Profit IR System for the Islamic Domain Leveraging Multiprocessing Design in Rust}

\author{Vera Pavlova \\
  rttl labs, UAE \\
  \texttt{v@rttl.ai} \\\And
  Mohammed Makhlouf\\
  rttl labs, UAE \\
  \texttt{mm@rttl.ai}\\}

\begin{document}
\maketitle
\begin{abstract}
The widespread use of large language models (LLMs) has dramatically improved many applications of Natural Language Processing (NLP), including Information Retrieval (IR). 
However, domains that are not driven by commercial interest often lag behind in benefiting from AI-powered solutions. One such area is religious and heritage corpora. Alongside similar domains, Islamic literature holds significant cultural value and is regularly utilized by scholars and the general public. Navigating this extensive amount of text is challenging, and there is currently no unified resource that allows for easy searching of this data using advanced AI tools. 
This work focuses on the development of a multilingual non-profit IR system for the Islamic domain. This process brings a few major challenges, such as preparing multilingual domain-specific corpora when data is limited in certain languages, deploying a model on resource-constrained devices, and enabling fast search on a limited budget. By employing methods like continued pre-training for domain adaptation and language reduction to decrease model size, a lightweight multilingual retrieval model was prepared, demonstrating superior performance compared to larger models pre-trained on general domain data. Furthermore, evaluating the proposed architecture that utilizes Rust Language capabilities shows the possibility of implementing efficient semantic search in a low-resource setting.

\end{abstract}

\section{Introduction}

Dense retrieval is an advanced approach in IR that utilizes embeddings to identify semantically similar text, known as semantic search. LLMs are a key component in creating text embeddings and performing dense retrieval \citep{karpukhin-etal-2020-dense, Izacard-unsupervised-dense}.
One of the first challenges in building a non-profit multilingual domain-specific IR system is that the use of publicly available multilingual large language models (MLLMs) pre-trained on a general domain could deteriorate performance due to domain shift when applied to new domains \citep{Lee-2019, Huang-ClinicalBERT}. To overcome this, we begin with pre-training an MLLM for the Islamic domain to address this issue. However, pre-training a domain-specific MLLM brings two additional challenges. Firstly, assembling a multilingual domain-specific corpus for pre-training a MLLM requires a large amount of domain-specific data that is often difficult to find in different languages. Secondly, multilingual models are heavyweight, frequently exceeding 1GB, making them challenging to deploy. To effectively tackle the issue of pre-training domain-specific MLLM, we employ a continued pre-training approach and incorporate domain-specific vocabulary to accommodate the domain shift better \citep{beltagy-etal-2019-scibert}. To deal with the challenge of the large size of MLLM, we perform language reduction and remove languages not needed in the current deployment. This method helps us reduce the model's size by more than half, even after introducing new domain-specific vocabulary.
We use this lightweight domain-specific MLLM as a backbone for the retrieval. Evaluation of this model on an in-domain IR dataset found that our model significantly outperforms general-domain multilingual and monolingual models even after performing language reduction.

\begin{figure*}[t]
  \includegraphics[width=\textwidth]{{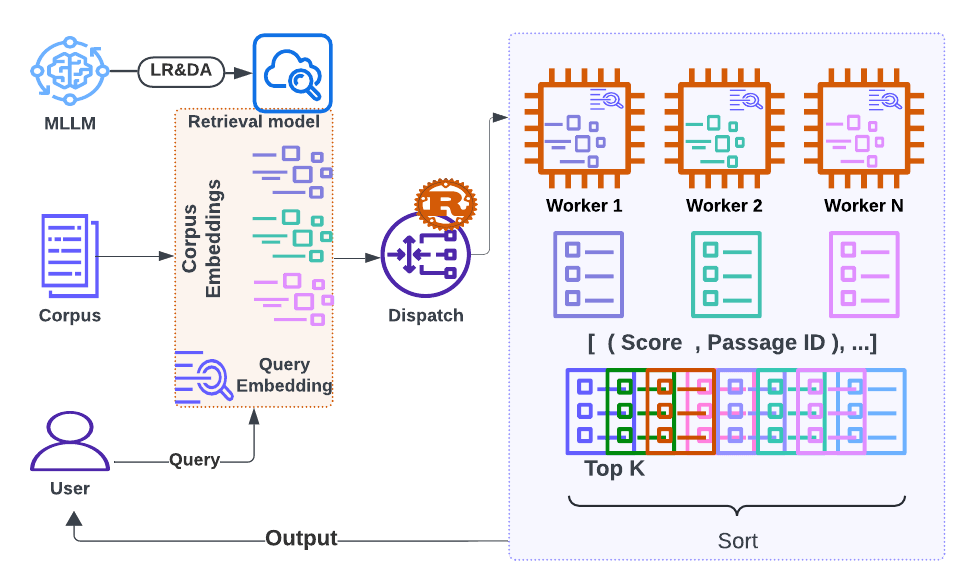}}
  \caption{The main components of building a multilingual IR system. In the upper left corner is the preparation of the retrieval model that includes language reduction (LR) and domain adaptation (DA). The rest of the figure shows the implementation of semantics search in Rust with multiprocessing architecture.}
  \label{fig:figure 1}
\end{figure*}

Moreover, deploying non-profit AI systems implies operating on a limited budget, which makes it challenging to use embedding APIs or libraries that rely on GPU acceleration to perform search reasonably fast. To tackle this challenge and meet the requirements of implementing an ad hoc IR system on a public website, we utilize the multiprocessing capabilities of Rust Language to create an efficient and secure semantic search based on CPU architecture \citep{Abdi-Fearless, Seidel-rust-safety, Liang-garbagecollection}.
Our system's evaluation and comparison against others, such as Faiss, indicates that our implementation of semantic search with underlying Rust multiprocessing architecture can significantly accelerate search without compromising performance.

Our main contributions are:
\begin{itemize}
    \item We have developed a free online multilingual search tool for exploring well-established literature in the Islamic domain.\footnote{A system is deployed at \url{https://rttl.ai/}}
    \item To the best of our knowledge, we are the first to deploy open-source, non-profit semantic search leveraging multiprocessing using Rust language.
\end{itemize}

\section{Lightweight Domain-Specific MLLM}

\subsection{Size Reduction of MLLM}
\begin{figure}[t]
  \includegraphics[width=\columnwidth]{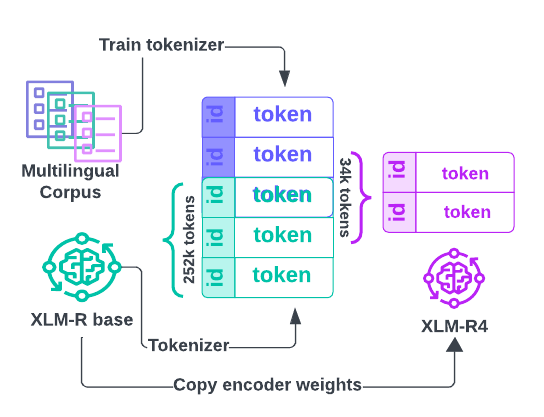}
  \caption{Language Reduction technique that gives us the multilingual XLM-R4 model for four languages (English, Russian, Arabic, and Urdu).}
  \label{fig:figure 2}
\end{figure}

MLLMs allow access to functionality in several languages using one model and enabling cross-lingual transfer. Pre-training mBERT
\citep{devlin-etal-2019-bert} and XLM \citep{Lample-XLM} on Wikipedia brought a new state-of-the-art to multilingual tasks.
\citet{conneau-etal-2020-unsupervised} showed that increasing MLLM's capacity and training on a larger corpus like CommonCrawl resulted in better-performing models such as XLM-R and XLM-R\textsubscript{Base}.
However, improved performance comes at the cost of the model's larger size (714MB for mBERT vs. 1.1GB for XLM-R\textsubscript{Base}).
The size of the model makes it heavy to deploy in low-resource settings. \citet{sun-etal-2019-patient, Tang-distillation, Sanh-distillation,  Li-distillation} showed that distillation of  transformer-based language models \citep{Vaswani-transformers} leads to considerable size reduction and adequate performance. Another approach that reduces model size and retains high performance is language reduction of  MLLM \citep{abdaoui-etal-2020-load}. Around 50\% of the parameters in mBERT and 70\% in XLM-R\textsubscript{Base} are assigned to the embedding matrix (see Table ~\ref{tab:Model size} in Appendix ~\ref{sec:appendix}).
Thus, applying language reduction is more favorable in the case of deploying MLLM as it decreases the model size while preserving encoder weights, trimming only the embedding matrix by removing the languages that are not needed in deployment.
Unlike \citet{abdaoui-etal-2020-load}, our reduction method involves training a new tokenizer (see Figure ~\ref{fig:figure 2}):

\begin{enumerate}
    \item We compile the corpus using a multilingual variant of the C4 corpus for the languages of interest (English, Russian, Arabic, and Urdu).
    \item Train the SentencePiece BPE tokenizer using this corpus.
    \item Find the intersection between the newly trained tokenizer and the original XLM-R\textsubscript{Base} tokenizer available from Hugging Face,\footnote{\url{https://huggingface.co/FacebookAI/xlm-roberta-base}} the tokens inside of intersection and corresponding weights will be selected for the new embedding matrix of the XLM-R4 model (34k tokens).
    \item We modify the SentencePiece model according to the new tokenizer.
    \item At the final stage, we copy the encoder weights from XLM-R\textsubscript{Base} to the new XLM-R4 model.
\end{enumerate}

The main difference in parameter size between the mBERT and XLM-R\textsubscript{Base} model is in the size of the embedding matrix (mBERT has 119K tokens, while the XLM-R\textsubscript{Base} has 252K tokens), while the size of encoder parameters of mBERT and XLM-R\textsubscript{Base} are the same. By only reducing the size of the embedding matrix of the XLM-R\textsubscript{Base}, we can significantly decrease the model's size to the size of the bert model or even smaller while benefiting from the extensive training that the XLM-R\textsubscript{Base} model underwent. The resulting XLM-R4 model, with a size of 481 MB and 119M parameters, is significantly smaller than XLM-R\textsubscript{Base}, demonstrating the practical implications of our method and its potential for real-world applications (see Table ~\ref{tab:Model size}). 

Table ~\ref{tab:XNLI comparison} compares how the models perform on the XNLI dataset \citep{conneau-etal-2018-xnli} in the cross-lingual transfer (fine-tuning multilingual model on English training set). As a baseline model, we use an XLM-R\textsubscript{Base}. Hugging Face implementation of the tokenizer of XLM-R\textsubscript{Base} is different from the original implementation \citep{conneau-etal-2020-unsupervised}. For a fair comparison, we fine-tune the XLM-R\textsubscript{Base} and the XLM-R4 model with the same hyperparameters on the English training set of the XNLI dataset (see Appendix~\ref{sec:appendix}). We also include in comparison mBERT, DistilmBERT \citep{Sanh-distillation}, and a reduced version of mBERT that consists of 15 languages \citep{abdaoui-etal-2020-load}. We compare the four languages left after performing the language reduction technique (English, Russian, Arabic, Urdu). Table ~\ref{tab:XNLI comparison} shows that the best-performing model for all languages is the XLM-R\textsubscript{Base} (in bold), and the second best-performing model (underlined) is the XLM-R4. We can observe a slight drop in performance of the XLM-R4 in comparison to the XLM-R\textsubscript{Base}, which is the smallest for Urdu (0.5\%) and English and Arabic (1.16\% and 1.65\% correspondingly), with a more noticeable drop in Russian (3.76\%). However, XLM-R4 performs better than the rest of the models, including mBERT. DistilmBERT shows the lowest results in all languages.
\begin{table}[]
\begin{adjustbox}{width=\columnwidth,center}
\begin{tabular}{ccccc}
\hline
\textbf{Model}        & \textbf{en}             & \textbf{ru}             & \textbf{ar}             & \textbf{ur}             \\ \hline
XLM-R\textsubscript{Base}    & \textbf{84.19} & \textbf{75.59} & \textbf{71.66} & \textbf{65.27} \\
XLM-R4       & {\ul 83.21}    & {\ul 72.75}    & {\ul 70.48}    & {\ul 64.95}    \\
mBERT        & 82.1           & 68.4           & 64.5           & 57             \\
mBERT 15lang & 82.2           & 68.7           & 64.9           & 57.1           \\
DistillmBERT & 78.5           & 63.9           & 58.6           & 53.3           \\ \hline
\end{tabular}
\end{adjustbox}
\caption{Results on cross-lingual transfer for four languages of the XNLI dataset. XLM-R\textsubscript{Base} and XLM-R4 results are averaged over five different seeds.}
\label{tab:XNLI comparison}
\end{table}

\begin{table}[]
\begin{adjustbox}{width=\columnwidth,center}
\begin{tabular}{cccc}
\hline
Model     & Size   & \#params & EM    \\ \hline
mBERT     & 714 MB & 178 M    & 92 M  \\
XLM-R\textsubscript{Base} & 1.1 GB & 278 M    & 192 M \\
XLM-R4    & 481 MB & 119 M    & 33M   \\ \hline
\end{tabular}
\end{adjustbox}
\caption{Comparison of models' size}
\label{tab:Model size}
\end{table}

\subsection{Domain Adaptation of MLLM}
\begin{figure}[t]
  \includegraphics[width=\columnwidth]{{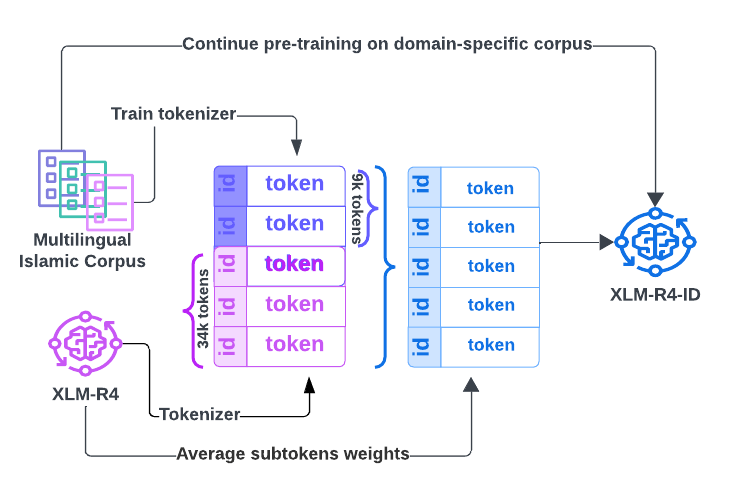}}
  \caption{Domain Adaptation of XLM-R4 utilizing continued pre-training approach on Multilingual Islamic Corpus. The final domain-specific model is XLM-R4-ID.}
  \label{fig:figure 3}
\end{figure}

The XLM-R\textsubscript{Base} model on which we perform language reduction to get the XLM-R4 model is pre-trained on the general domain. We perform domain adaptation of XLM-R4 to account for the domain shift  \citep{Lee-2019, Huang-ClinicalBERT}. 
One of the challenges here is the preparation of a multilingual Islamic corpus to adapt the XLM-R4 to the Islamic domain. 
The situation regarding constructing a multilingual corpus in the Islamic Domain is unusual. In most multilingual corpora, the data is predominantly in English, but in the Islamic domain, it is predominantly in Arabic. The Open Islamicate Texts Initiative (OpenITI) \citep{romanov2019openiti} has provided a sizable corpus (1 billion words) for pre-training LLMs in Classical Arabic, which is the language of Arabic Islamic literature. For English, Russian, and Urdu (50 million words altogether), the available text mainly consists of Tafseer (Qur’an exegesis) and Hadith. To avoid having a corpus heavily skewed towards Arabic, we selected a random subset of the OpenITI corpus containing approximately 250 million words. We combine it with content from other languages, resulting in a corpus of size 300M words for domain adaptation.
The corpus size is relatively small; nevertheless, since the weights of the XLM-R4 model are not initialized from scratch, we can apply continued pre-training. To address domain shift more effectively, we introduce new domain-specific vocabulary \citep{Gu-PubMedBERT, beltagy-etal-2019-scibert, Poerner-etal-2020-inexpensive, pavlova-makhlouf-2023-bioptimus}. The domain adaptation of XLM-R4 involves the following steps  (see Figure ~\ref{fig:figure 3}):

\begin{enumerate}
    \item We train a new SentencePiece BPE tokenizer using a multilingual Islamic Corpus. 
    \item We find the intersection between the new Islamic tokenizer and the XLM-R4 tokenizer. All the tokens outside of the intersection (9k tokens) are added to the embedding matrix, and the weights for new tokens are assigned by averaging existing weights of subtokens from the XLM-R4 model. 
    \item We continue pre-training XLM-R4 using the domain-specific corpus mentioned above to get the XLM-R4-ID (Islamic domain) model. For more details on the hyperparameters, refer to Appendix~\ref{sec:appendix}.
\end{enumerate}

\begin{table*}[thb!]
\centering
\resizebox{\linewidth}{!}{\def\arraystretch{1.5}%
\begin{tabular}{ccccccccc}
\hline
                                 & \multicolumn{2}{c}{\textbf{EN}}                                             & \multicolumn{2}{c}{\textbf{AR}}                                             & \multicolumn{2}{c}{\textbf{RU}}                                             & \multicolumn{2}{c}{\textbf{UR}}                                             \\
\multirow{-2}{*}{\textbf{Model}} & \multicolumn{1}{l}{Recall@100}       & \multicolumn{1}{l}{MRR@10}           & \multicolumn{1}{l}{Recall@100}       & \multicolumn{1}{l}{MRR@10}           & \multicolumn{1}{l}{Recall@100}       & \multicolumn{1}{l}{MRR@10}           & \multicolumn{1}{l}{Recall@100}       & \multicolumn{1}{l}{MRR@10}           \\ \hline
XLM-R\textsubscript{Base} (en)                   & 18.7                                 & 34                                   & 2.94                                 & 6.94                                 & 17.9                                 & 31.8                                 & 20.4                                 & 33.7                                 \\
XLM-R\textsubscript{Base} (ar)                   & 17.8                                 & 32.9                                 & 5.3                                  & 6.3                                  & 20                                   & 30.1                                 & 20.7                                 & 33.9                                 \\
XLM-R4-ID (en)                   & 27.2                                 & 43.8                                 & 28.6                                 & {\color[HTML]{036400} \textbf{45.5}} & {\color[HTML]{6200C9} \textbf{24.5}} & 34.7                                 & 26.8                                 & 40                                   \\
XLM-R4-ID (ar)                   & {\color[HTML]{00009B} \textbf{27.8}} & {\color[HTML]{00009B} \textbf{45.5}} & {\color[HTML]{036400} \textbf{29.3}} & {\color[HTML]{036400} \textbf{45.5}} & 24.1                                 & {\color[HTML]{6200C9} \textbf{37.5}} & {\color[HTML]{F56B00} \textbf{27.3}} & {\color[HTML]{F56B00} \textbf{41.5}} \\
ST/multilingual-mpnet-base-v2    & 21.6                                 & 34.3                                 & 4.8                                  & 5.2                                  & 17.2                                 & 22.4                                 & 13.5                                 & 19.1                                 \\
ST/all-mpnet-base-v2             & 25                                   & 40.9                                 & -                                    & -                                    & -                                    & -                                    & -                                    & -                                    \\ \hline
\end{tabular}
}

\caption{Performance on in-domain IR dataset for four languages. The best scores are in bold, and color codes correspond to different languages.}
\label{tab:In-domain performance}
\end{table*}

\section{Domain-specific IR}
To prepare the retrieval model, we utilize a dense retrieval approach  \citep{karpukhin-etal-2020-dense} that employs dual-encoder architecture \citep{Bromley-Siamese}. We use the sentence transformer framework that adds a pooling layer on top of LLM embeddings and produces fixed-sized sentence embedding \citep{reimers-gurevych-2019-sentence}. The loss function is formulated in the framework of contrastive learning that enables learning an embedding space that brings closer queries and their relevant passages and pushes further queries and irrelevant passages \citep{Oord-Contrastive}. For efficient training, we use in-batch negatives \citep{henderson2017efficient, gillick-etal-2019-learning, karpukhin-etal-2020-dense}. 
The transfer language of the XLM-R\textsubscript{Base} is English, while XLM-R4-ID was adapted for the Islamic Domain, predominately using Arabic. We experiment with both English and Arabic as transfer languages to compare their transfer potential for solving the IR task at hand.
We utilize the MS MARCO IR dataset, which contains more than half a million queries and a collection of 8.8M passages in English \citep{Bajaj-MSMARCO} to allow cross-lingual transfer from English and we use an Arabic machine-translated version of MS MARCO \citep{Bonifacio-mmarco} employing Arabic as transfer language. 
Consequently, we prepared four retrieval models, training XLM-R\textsubscript{Base} and XLM-R4-ID, using English and Arabic MS MARCO (for hyperparameters details see Appendix~\ref{sec:appendix}).
For evaluation, we use Arabic QRCD (Qur'anic Reading Comprehension Dataset) \citep{malhas2020ayatec} as IR Dataset and its verified translation to English, Russian and Urdu. We use train and development sets (169 queries) for testing. As a collection for retrieval, we use the Holy Quran text (Arabic), Sahih International translation (English), 
Elmir Kuliev (Russian) and Ahmed Raza Khan (Urdu) are available on tanzil.net.\footnote{\url{https://tanzil.net/trans/}}
We evaluate the models' performance using Recall@100 and the order-aware metric MRR@10 (MS MARCO's official metric). 

In Table ~\ref{tab:In-domain performance}, we compare different models, including the SentenceTransformer model (paraphrase-multilingual-mpnet-base-v2), which was trained by distilling knowledge from the teacher model paraphrase-mpnet-base-v2 and using XLM-R\textsubscript{Base} as the student model. Additionally, we assess the performance of the monolingual teacher model paraphrase-mpnet-base-v2 in English.
The table shows that both XLM-R4-ID models outperform the others, including the monolingual model (ST/all-mpnet-base-v2). Even though XLM-R4 is a reduced version of XLM-R\textsubscript{Base}, it significantly outperforms XLM-R\textsubscript{Base}. This improvement in performance shows that domain adaptation was beneficial. It is also important to mention that both the XLM-R\textsubscript{Base} and the multilingual-mpnet-base-v2 models perform poorly in Arabic. This observation may indicate that domain shift might have a significant impact, particularly with the Arabic language.
Moreover, we observe that XLM-R4-ID trained on the Arabic machine-translated version of MS MARCO outperforms XLM-R4-ID trained on English MS MARCO for all languages with one exception of Recall@100 metric for Russian. These results can be explained by the fact that a significant part of the corpus for domain adaptation was in Arabic (around 85\%). We can suggest that Arabic can effectively function as a transfer language for the Islamic domain. For all subsequent sections of the paper and for deployment, we will be using XLM-R4-ID (ar).

\begin{table*}[]
\resizebox{\linewidth}{!}{\def\arraystretch{1.5}%
\begin{tabular}{cccccccccc}
\hline
\textbf{SUT}     & \textbf{\begin{tabular}[c]{@{}c@{}}Python\\ (e.s.)\end{tabular}} & \textbf{\begin{tabular}[c]{@{}c@{}}HNSW\end{tabular}} & \textbf{\begin{tabular}[c]{@{}c@{}}SQ\\ (e.s.)\end{tabular}} & \textbf{\begin{tabular}[c]{@{}c@{}}PQ\\ (e.s.)\end{tabular}} & \textbf{\begin{tabular}[c]{@{}c@{}}Rust \\ 1 w. (e.s.)\end{tabular}} & \textbf{\begin{tabular}[c]{@{}c@{}}Rust\\ 2 w. (e.s.)\end{tabular}} & \textbf{\begin{tabular}[c]{@{}c@{}}Rust\\ 4 w. (e.s.)\end{tabular}} & \textbf{\begin{tabular}[c]{@{}c@{}}Rust\\ 6 w. (e.s.)\end{tabular}} \\ \hline
\textbf{Speedup} & 1x                                                             & 5x                                                                                & 3.9x                                                                             & 9x                                                                                                           & 2.6x                                                                     & 3.8x                                                                     & 4.5x                                                                     & 4.9x                                                                     \\
\textbf{Recall}  & 100\%                                                                    & 90\%                                                                                & 90\%                                                                             & 85\%                                                                                                                             & 100\%                                                                    & 100\%                                                                    & 100\%                                                                    & 100\%                                                                    \\ \hline
\end{tabular}
}

\caption{Comparisons of SUTs for the speedup of retrieval against baseline and percentage of baseline Recall (e.s stands for exact search and w. for worker).}
\label{tab:SUT comparison}

\end{table*}

\section{Deploying Domain-Specific IR System}
Using GPUs to train transformer-based LLMs and retrieval models is often a necessity. However, GPUs for inference in a production environment 
are cost-prohibitive, especially in non-profit organizations.
Additionally, given supply availability to ensure the right size of cloud machines with GPUs often imposes a fixed set of resources in predefined bundles of size, which typically leads to vast overprovisioning and grossly underutilized resources.
Our goal is to maximize software performance and resource efficiency on widely-used, cost-effective CPU servers. We argue that leveraging the ubiquity and flexibility of CPU servers makes it possible to build a system and improve efficiency independently of the underlying substrate, allowing deployment even on serverless infrastructure, which is predominately CPU-based. 

\subsection{Rust for Production AI Workloads}
Production use of IR systems requires real-time processing capabilities.  
However, the main challenge of using state-of-the-art retrieval models in production is their high inference time. Deploying such models on resource-constrained devices is even more problematic. A few approaches like model quantization \citep{Guo-quantized, Jacob-quantization, bondarenko-etal-2021-understanding, tian-etal-2023-samp}, embedding size compression \citep{Zhu-pca, gupta-etal-2019-improving, Kusupati-matryoshka, li-espresso} can help to address this issue at the cost of model performance. However, in specific applications of semantic search, such as Islamic Domain, even a slight decrease in performance is highly undesirable. We argue that it is possible to improve inference times without compromising search quality. To minimize the trade-off between latency and performance, we leverage the advantages of the Rust language.

Rust is a safe and efficient systems programming language that addresses many pain points in other commonly used interpreted languages, such as Python, which imposes the presence of the Python interpreter in the production environment. Providing zero-cost abstractions to the hardware substrate with a lightweight memory footprint, idiomatically written Rust outperforms identical equivalents written in JVM-based languages such as Java \citep{Perke-rust-scientists}.
The absence of garbage collection mechanics in Rust makes systems written in Rust more deterministic and better suited for production deployments in serverless and compact runtimes where compute is billed by milliseconds \citep{Liang-garbagecollection}.
The borrow checker of Rust eliminates an entire class of security vulnerabilities introduced by references outliving the data they point to. This feature guarantees safety, especially when writing concurrent and multiprocessing code, without sacrificing performance gains \citep{Seidel-rust-safety, Jung-safe-rust, Abdi-Fearless}.
Energy efficiency and reduced carbon footprint are other crucial features of using Rust in AI production workloads \citep{Pereira-energy}.

\begin{figure}[t]
  \includegraphics[width=\columnwidth]{{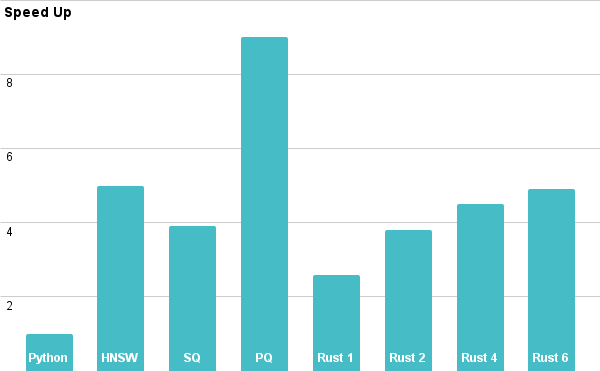}}
  \caption{Speedup and Recall of SUTs.}
  \includegraphics[width=\columnwidth]{{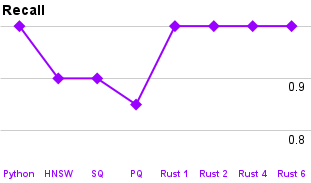}}
  \caption{Speedup and Recall of SUTs.}
  \label{fig:figure 4}
\end{figure}

\subsection{System Design for Rust-based Semantic Search}
Such libraries as Faiss\footnote{\url{https://ai.meta.com/tools/faiss/}} offer the best speedup using GPU architecture, which significantly increases deployment costs. Faiss also provides multi-threading capabilities but lacks native cost-efficient multiprocessing and true parallelism for individual search queries. The best CPU performance is achieved by sending queries in batches, which does not align with real-world online search.
Utilizing the Rust language's capabilities enables us to implement a multiprocessing architecture efficiently and securely for our IR system. We built the system on top of the Candle framework,\footnote{\url{https://github.com/huggingface/candle}} a minimalist machine-learning framework for Rust. The system's architectural design goes as follows (see Figure ~\ref{fig:figure 1}):

\begin{enumerate}
    \item The passages from the corpus are converted to embeddings and stored for caching during the search.
    \item The corpus embeddings are divided into chunks and distributed across the specified number of workers.
    \item For multiprocessing during the search, an embedding of a search query is sent to each worker asynchronously.
    \item Each worker conducts an exact search by comparing the query with each passage within the allocated chunk and then assigns a score using the similarity function.
    \item The workers then return scores to the main thread as a list of tuples, each containing a score and a passage ID for sorting.
    \item At the final stage, the scores are sorted in descending order, and the corresponding passages are returned to the user based on the topk parameter.
\end{enumerate}

We compare our system's performance against Faiss 
implementation of the following algorithms: Hierarchical Navigable Small World graph (HNSW), Scalar Quantization with fp16 (SQ), and Product Quantization (PQ). 
To compare the Systems Under Test (SUTs), we assume the following conditions: the corpus and query embeddings are precomputed and preloaded in memory. To accommodate different minute fluctuations posed by potential hardware condition variance, we average ten runs of each system across test queries that are provided linearly. All the systems perform the search and retrieval using CPU-based architecture. To test on a bigger retrieval corpus (approx 50k passages), the dataset used for measuring the time of retrieval and Recall of SUTs is Hadith Question-Answer pairs (HAQA) \citep{alnefaie-etal-2023-haqa}. The similarity function utilized during the search is cosine similarity. All the systems employ XLM-R4-ID (ar) as a retrieval model. The hardware used for the test is a cloud instance (1x NVIDIA A10) provided by Lambda Labs's public cloud.

Table ~\ref{tab:SUT comparison} highlights the trade-off between retrieval time and performance for different SUTs. The main focus of comparison is the speed of retrieval. Python implementation of exact search is a baseline with its score for Recall@100 (Recall) taken as 100\%. We can observe that the speedup of retrieval time of Faiss algorithms always comes at the cost of lower Recall. At the same time, the implementation of semantic search in Rust doesn't endure the trade-off between retrieval time and performance. Figure ~\ref{fig:figure 4} illustrates the dip in Recall plot for the highest speedup of the PQ algorithm while Recall for Rust implementation stays flat at 100\% for all instances. Moreover, a speedup of 2.6 times is achievable with Rust implementation without applying multiprocessing (using one worker), and further speedup is possible by adding more workers.

\section{Related work}
There is a substantial amount of work written on the topic of pre-training domain-specific LLM; some of them describe more costly approaches like pre-training a new LLM from scratch \citet{Gu-PubMedBERT, beltagy-etal-2019-scibert}, some more resource-efficient approaches like continued pre-training \citet{Lee-2019, Huang-ClinicalBERT}, and there is a body of work that research methods of domain-adaptation in a low resource setting \citet{Poerner-etal-2020-inexpensive, sachidananda-etal-2021-efficient, pavlova-2023-leveraging}.
The survey \citet {Zhao-dense-survey} covers in detail the topic of dense retrieval, discussing different types of models' architecture and training approaches, including the selection of high-quality negatives.
There is a growing body of research on Rust Language memory-safe features that came to be known as fearless concurrency \citep{Jung-safe-rust, Abdi-Fearless, Evans-unsafe-rust, Perke-rust-scientists}. 

\section{Conclusion}

This work outlines the development of a non-profit multilingual IR system for the Islamic domain. We also address the challenges it presents and propose potential solutions for handling these challenges in low-resource settings. Our research demonstrates that utilizing continued pre-training and integrating new domain-specific vocabulary can help mitigate domain shift, even when pre-training on a small corpus. The retrieval model we built using a domain-adapted MLLM as a foundation exhibited better performance compared to general domain models. Additionally, we found that implementing language reduction can significantly decrease the model size without deteriorating performance.
Furthermore, we showed that leveraging the multiprocessing capabilities of the Rust language can decrease inference time without compromising performance or requiring expensive acceleration hardware like GPUs.

\section*{Limitations}
 To measure the inference time and recall of SUTs we are restricted to using a smaller retrieval corpus (around 50k passages). The real size of the data for retrieval is above 150k passages.

\section*{Acknowledgment}
Developing a multiprocessing CPU-based search with Rust would not have been possible without Mohamed Samir from SYWA AI. We would also like to express our gratitude to Osama Khalid from SYWA AI for assisting in verifying the quality of the Urdu translation of the QRCD queries. We extend our thanks to the anonymous Reviewers and the Area Chair for their valuable feedback and to the Program Chairs for promptly addressing and resolving all related matters.

\bibliography{anthology,custom}

\appendix
\section{Appendix}
\label{sec:appendix}

\begin{table}[h]
\begin{adjustbox}{width=\columnwidth,center}
\begin{tabular}{ccc}
\hline
\textbf{Computing Infrastructure}    & \multicolumn{2}{c}{1x H100 (80 GB)}    \\ \hline
\multicolumn{2}{c}{\textbf{Hyperparameter}} & \textbf{Assignment}                     \\ \hline
\multicolumn{2}{c}{number of epochs}        & 60 \\ \hline
\multicolumn{2}{c}{batch size}              & 128                             \\ \hline
\multicolumn{2}{c}{maximum learning rate}   & 0.0005                        \\ \hline
\multicolumn{2}{c}{learning rate optimizer} & Adam                                    \\ \hline
\multicolumn{2}{c}{learning rate scheduler} & None or Warmup linear                   \\ \hline
\multicolumn{2}{c}{Weight decay}            & 0.01                                    \\ \hline
\multicolumn{2}{c}{Warmup proportion}       & 0.06                                    \\ \hline
\multicolumn{2}{c}{learning rate decay}     & linear                                  \\ \hline
\end{tabular}
\end{adjustbox}
\caption{Hyperparameters for pre-training of XLM-R4-ID model.}
\label{tab:appendix-table-a}
\end{table}

\begin{table}[!htbp]
\begin{adjustbox}{width=\columnwidth,center}
\begin{tabular}{ccc}
\hline
\textbf{Computing Infrastructure} & \multicolumn{2}{c}{2 x NVIDIA RTX 3090 GPU} \\ \hline

\multicolumn{2}{c}{\textbf{Hyperparameter}}     & \textbf{Assignment}           \\ \hline
\multicolumn{2}{c}{number of epochs}            & 10                            \\ \hline
\multicolumn{2}{c}{batch size}                  & 8                             \\ \hline
\multicolumn{2}{c}{learning rate}               & 2e-5                          \\ \hline
\multicolumn{2}{c}{weight decay}                     & 0.01                         \\ \hline

\end{tabular}
\end{adjustbox}
\caption{Hyperparameters for fine-tuning on XNLI dataset.}
\label{tab:appendix-table-b}
\end{table}

\begin{table}[!htbp]
\begin{adjustbox}{width=\columnwidth,center}
\begin{tabular}{ccc}
\hline
\textbf{Computing Infrastructure} & \multicolumn{2}{c}{1x H100 (80 GB)} \\ \hline

\multicolumn{2}{c}{\textbf{Hyperparameter}}     & \textbf{Assignment}           \\ \hline
\multicolumn{2}{c}{number of epochs}            & 10                            \\ \hline
\multicolumn{2}{c}{batch size}                  & 256                             \\ \hline
\multicolumn{2}{c}{learning rate}               & 2e-5                          \\ \hline
\multicolumn{2}{c}{pooling}                     & mean                         \\ \hline

\end{tabular}
\end{adjustbox}
\caption{Hyperparameters for training retrieval models.}
\label{tab:appendix-table-c}
\end{table}

\end{document}